\documentclass[runningheads]{llncs}
\usepackage{graphicx}
\usepackage{amsfonts}

\usepackage[misc]{ifsym}
\usepackage{booktabs}
\usepackage{multirow}
\usepackage[linesnumbered, ruled]{algorithm2e}
\usepackage[colorlinks,
            linkcolor=blue,       
            anchorcolor=blue,  
            citecolor=blue,
            urlcolor=blue
            ]{hyperref}

\usepackage{tikz,xcolor,hyperref}
 
\definecolor{lime}{HTML}{A6CE39}
\DeclareRobustCommand{\orcidicon}{%
	\begin{tikzpicture}
	\draw[lime, fill=lime] (0,0) 
	circle [radius=0.16] 
	node[white] {{\fontfamily{qag}\selectfont \tiny ID}};
	\draw[white, fill=white] (-0.0625,0.095) 
	circle [radius=0.007];
	\end{tikzpicture}
	\hspace{-2mm}
}
 
\foreach \x in {A, ..., Z}{%
	\expandafter\xdef\csname orcid\x\endcsname{\noexpand\href{https://orcid.org/\csname orcidauthor\x\endcsname}{\noexpand\orcidicon}}
}

\begin{document}
\title{Evolutionary Verbalizer Search for Prompt-based Few Shot Text Classification}
\titlerunning{Evolutionary Verbalizer Search}
%
\author{Tongtao Ling \and Lei Chen\textsuperscript{(\Letter)}\orcidA{} \and Yutao Lai \and Hai-Lin Liu\orcidB{}}

%
\authorrunning{T. Ling et al.}
%

\institute{Guangdong University of Technology, Guangzhou, China \\
\email{ltt\_rick@163.com, chenlei3@gdut.edu.cn, tg980515@163.com, hlliu@gdut.edu.cn}}
\maketitle              
\begin{abstract}
Recent advances for few-shot text classification aim to wrap textual inputs with task-specific prompts to cloze questions. By processing them with a masked language model to predict the masked tokens and using a verbalizer that constructs the mapping between predicted words and target labels. This approach of using pre-trained language models is called prompt-based tuning, which could remarkably outperform conventional fine-tuning approach in the low-data scenario. As the core of prompt-based tuning, the verbalizer is usually handcrafted with human efforts or suboptimally searched by gradient descent. In this paper, we focus on automatically constructing the optimal verbalizer and propose a novel evolutionary verbalizer search (EVS) algorithm, to improve prompt-based tuning with the high-performance verbalizer. Specifically, inspired by evolutionary algorithm (EA), we utilize it to automatically evolve various verbalizers during the evolutionary procedure and select the best one after several iterations. Extensive few-shot experiments on five text classification datasets show the effectiveness of our method.   

\end{abstract}
\section{Introduction}

In recent years, pre-trained language models (PLMs)~\cite{devlin2018bert} have achieved great success in NLP tasks such as natural language understanding and natural language generation. 
The conventional method of utilizing PLMs for downstream tasks is fine-tuning, where we add a classifier to the top of PLMs and further train on sufficient labeled data~\cite{howard2018universal}. Fine-tuning achieves satisfactory results on supervised downstream tasks. However, since the additional classifier requires sufficient training instances for fitting, it can not replicate the same success in low-resource scenarios (zero and few-shot tasks). In addition, training the additional classifier is usually difficult due to the gap between pre-trained tasks (e.g., masked language modeling) and downstream tasks (e.g., text classification). To this end, prompt-tuning, a novel paradigm of NLP has risen to be a powerful way for low-resource works to narrow the gap between pre-trained stage and downstream tasks stage~\cite{schick2020exploiting}. 

The core of prompt-tuning is to re-formalize a classification task to a cloze-style task. For example, in a sentiment classification task, given an original input sentence ``I like eating apples'', we add a task-specific prompt (also called template) ``It is {\tt [MASK]}'' after the sentence, then the PLMs will predict the token {\tt [MASK]} token and the token is further mapped to the target label by a \textbf{verbalizer} (e.g., ``great'' to ``positive''). Verbalizer can be seen as a mapping function from label word (label word means the set of predicted words corresponding labels) space to label space. It is the connection between {\tt [MASK]} token outputs and final predicted results~\cite{gao2020making}. Since the verbalizer can directly determine the effectiveness of the classification result, how to construct the verbalizer is a critical issue in prompt-tuning~\cite{liu2021pre}. 

\begin{figure}[t]
    \centering    
     \includegraphics[width=0.8\linewidth]{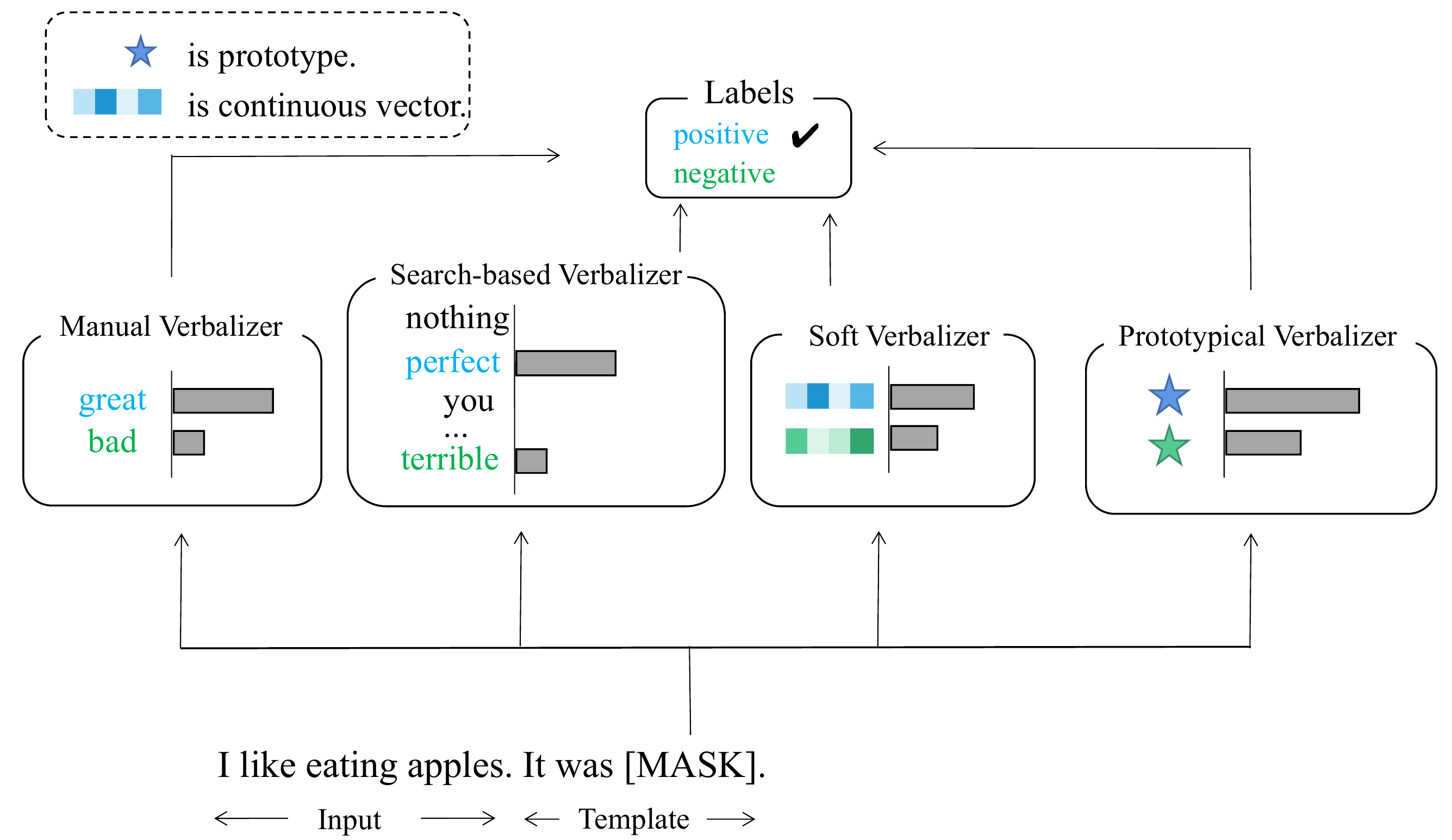}
    \caption{Illustration of current verbalizer construction methods.}
    \label{fig:fig1}
\end{figure}

Recently, the construction of verbalizer is mainly divided into the following categories: manual verbalizer, search-based verbalizer, soft verbalizer, and prototypical verbalizer. We show an illustration of them in Fig.\ref{fig:fig1}. The most commonly used verbalizer is the manual verbalizer, which can be divided into two categories: One2One and One2Many. One2One verbalizer means that each label only has one label word, which requires human effort and domain knowledge to select the best label word. But One2One limits the coverage of label words because it cannot summarize the semantic information of the label using one label word. Therefore, to enrich the semantic information of label words for a specific label,~\cite{hu2021knowledgeable} proposes to select some related words from a large knowledge base to construct the One2Many verbalizer, which has multiple label words for each label. These approaches can greatly improve the semantics of labels, but related words of the different label may overlap, which expect human efforts to filter some overlapped words and select the remaining suitable words to construct a task-specific verbalizer. To solve these issues, search-based verbalizer aims to find suitable label words with various algorithms~\cite{schick2020automatically} and soft verbalizer treats label words as trainable tokens which are optimized in the training process~\cite{hambardzumyan2021warp}. However, when the amount of annotated data is not sufficient, the search-based is not necessarily better than the manual verbalizer.
To this end, the prototypical verbalizer aims to obtain a semantic representation of labels and project the hidden vectors of {\tt [MASK]} to the embedding space and learn prototypes with few data for each label to serve as a verbalizer. Despite these efforts, the challenge of obtaining the high-performance verbalizer under low-resource scenario for prompt-tuning still exists, and previous works pointed out that the current gradient-based verbalizer search is usually suboptimal~\cite{liu2021pre}. 

It is challenging to automatically find the optimal verbalizer for prompt-tuning. Inspired by the Evolutionary Algorithm (EA)~\cite{blickle1996comparison}, we propose a novel search-based verbalizer approach for this purpose. Evolutionary algorithms are a kind of metaheuristic method inspired by natural population evolution. Compared to the traditional gradient-based search, EA can conduct a more robust and global search by leveraging the internal parallelism of the search population. To search for optimal label words as verbalizer, in this paper, we introduce a novel Evolutionary Verbalizer Search (EVS) algorithm to find the optimal One2Many verbalizer for text classification on a tiny set of labeled data. EVS does not require updating any model parameters because it relies on a small development set for validation rather than training. To verify the effectiveness of the proposed EVC, extensive experimental studies are conducted on benchmark datasets, and the experimental results are compared with various verbalizes. Our experiments show that EVS can achieve substantial improvements over the baseline of manual verbalizer and other search-based verbalizers, and it also outperformed the other state-of-the-art verbalizers.

In summary, the main contributions of this paper are as follows:
\begin{itemize}
    \item We design a novel evolutionary verbalizer search algorithm that can automatically find the high-performance verbalizer on a small development set.
    \item For the few-shot scenario, we conduct experiments with other state-of-the-art verbalizers, and our proposed methods outperform other existing search-based verbalizers in the same few-shot setting and achieve similar performance with manual verbalizers. 
    \item We carefully investigate the effects of hyperparameters in the proposed method. And we also find that our searched verbalizers are semantically related to the corresponding class.
\end{itemize}

\section{Related Work}
In this section, the related work to this study is introduced briefly.
\subsection{Verbalizer Construction}

In this paper, we investigate the search-based method for verbalizer construction in prompt-tuning. As a crucial component of prompt-tuning, verbalizer aims to transform model outputs to target labels and has a strong influence on the performance of prompt-tuning~\cite{gao2020making}.  Most previous works focused on manual verbalizers, which require manual elaboration and enough human prior knowledge. To avoid the drawbacks of manual verbalizers, some studies propose to construct an automatic verbalizer by using search-based methods~\cite{schick2020automatically}, which are expected to find suitable label words during the training process. In addition to using discrete words to construct verbalizers, soft verbalizers have been proposed by~\cite{hambardzumyan2021warp}, which treat labels as trainable tokens and optimize in the training process. Some other works design prototypical verbalizers~\cite{wei2022eliciting,cui2022prototypical}, which learn prototype vectors as verbalizers by contrastive learning.
\subsection{Evolutionary Algorithm}

Due to the properties of gradient-free and highly parallelism, Evolutionary Algorithm (EA)~\cite{blickle1996comparison} has been applied to extensive applications. For instance, automated machine learning (AutoML), a typical combination of deep learning techniques and evolutionary algorithms, has demonstrated promising results in the areas of parameter optimization and neural architecture search (NAS)~\cite{elsken2019neural}. More than that, EAs have been applied to NLP tasks successfully. ~\cite{manzoni2020towards} use classical NLP methods incorporated with Genetic Programming (GP) to predict the next words with an input word. Furthermore, the complementary relationship between large language models and evolutionary computing is discussed in~\cite{lehman2022evolution}. However, this field is still worth exploring like how to leverage the benefits of EA for NLP tasks effectively.  

 \begin{figure}[t]
    \centering    
    \includegraphics[width=0.8\linewidth]{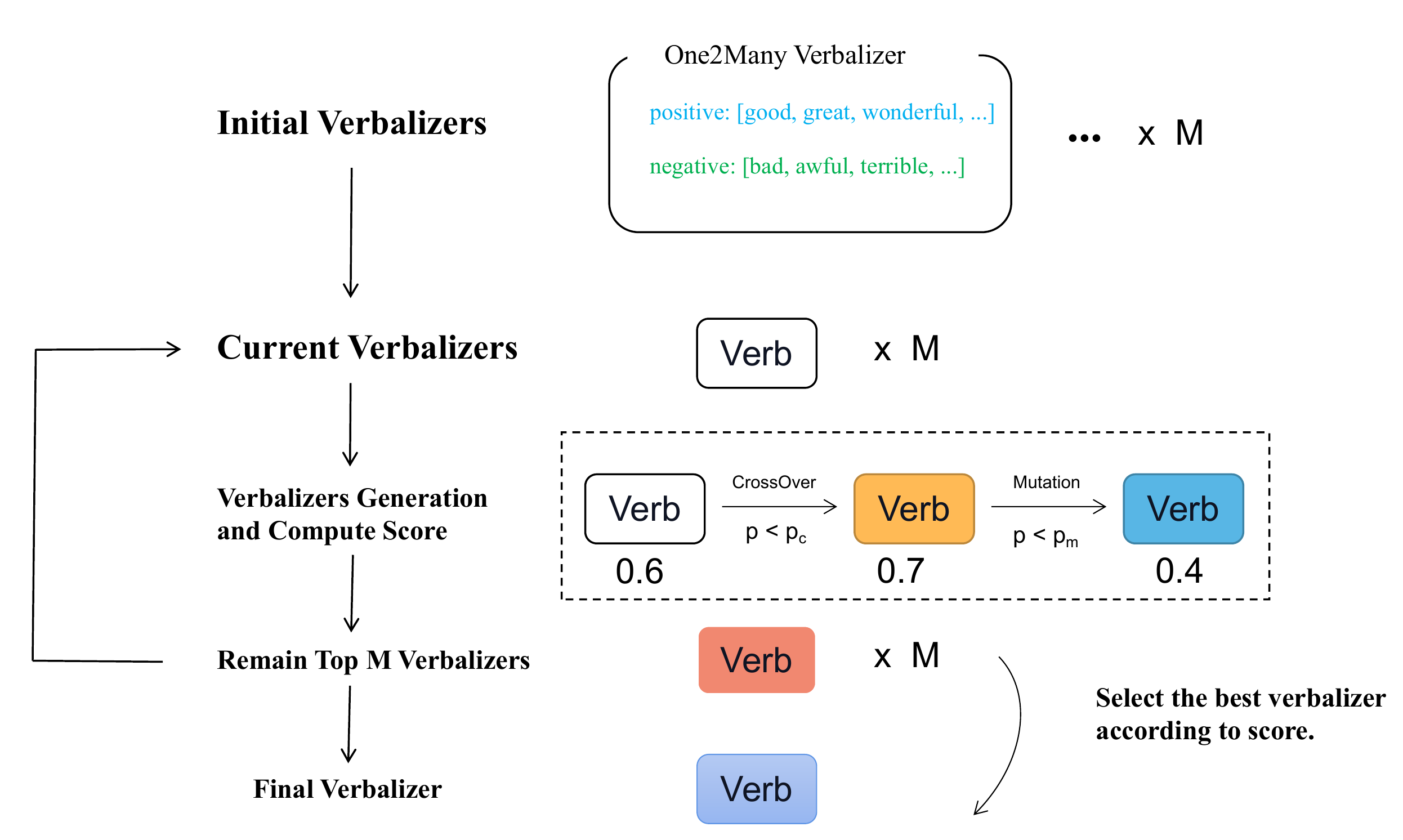}
    \caption{Illustrative procedure of EVS algorithm. The core of EVS is borrowed from evolutionary algorithm.}
    \label{fig:fig2}
\end{figure}

\section{Evolutionary Verbalizer Search}

In this section, the core idea of the proposed Evolutionary Verbalizer Search (EVS) is introduced first, and then the verbalizer generation strategies are given in detail. 

\subsection{Encoding and Decoding}
To conduct an evolutionary verbalizer search, proper encoding and decoding methods are first needed. Formally, denote $\mathcal{M}$, $\mathcal{T}$ as a masked language model (MLM) and a task-specific template function, respectively. We first sample a tiny labeled data as development set $D_{dev}$ and each instance $x$ in $D_{dev}$ is wrapped by $\mathcal{T}$ with a {\tt [MASK]} token, the prompt input can be formulated as:
\begin{equation}
    x_{t} = \mathcal{T}(x)
    \label{eq:eq1}
\end{equation}

Then, we input $x_{t}$ into $\mathcal{M}$ to compute the hidden vectors of {\tt [MASK]} token $h_{[mask]} \in \mathbb{R}^{V}$, which represents the probability distribution of the whole vocabulary. It can be calculated as:
\begin{equation}
    h_{[mask]} = \mathcal{M}(x_{t})
    \label{eq:eq2}
\end{equation}
where $V$ is the vocabulary size. In EVS, we input the whole $D_{dev}$ to $\mathcal{M}$ and compute each label average representation $h_{[mask]}^{i} \in \mathbb{R}^{V} $, $i \in \{label_{1},...,label_{N}\}$, $N$ is the number of labels. So we can get the initialized matrix $h_{init} \in \mathbb{R}^{N \times V }$. After that, let $N_{c}$ be the number of candidates for each label, which means that we need to select label words from candidates. Then, for each $label_{i}$, $i \in N$, we sort the elements of $h_{[mask]}^{i}$ and select top $N_{c}$ candidates to acquire $h_{c}^{i} \in \mathbb{R}^{N{c}}$. The encoding matrix can be computed as follows:
\begin{equation}
  h_{encode} =
  \left[ {\begin{array}{c}
    h_{c}^{1} \\
    \vdots \\
    h_{c}^{N} \\
  \end{array} } \right],   h_{encode} \in \mathbb{R}^{ N \times N_{c}}
  \label{eq:eq3}
\end{equation}

After that, we use a square matrix $X^{*} \in \mathbb{R}^{ N_{c} \times N_{c}}$ to multiply $h_{encode}$ as decoding:
\begin{equation}
  h_{decode} = h_{encode} \cdot X^{*}, h_{decode} \in \mathbb{R}^{ N \times N_{c}}
  \label{eq:eq4}
\end{equation}

Finally, let $N_{l}$ be the number of label words for each label, which represents the number of label words we expect to choose. And we also sort the elements of $h_{c}^{i}$ and memorize the index of vocabulary, then we select top $N_{l}$ elements to acquire $\hat{h_{c}^{i}} \in \mathbb{R}^{N_{l}}$ and change it into corresponding words by indexes. The decoding verbalizer is defined as: 
\begin{equation}
  h^{*} = \left[ {\begin{array}{c}
    \hat{h_{c}^{1}} \\
    \vdots \\
    \hat{h_{c}^{N}} \\
  \end{array} } \right] = 
  \left[ {\begin{array}{ccc}
    w_{11} & \cdots & w_{1N{l}} \\
    \vdots & \ddots & \vdots \\
    w_{N1} & \cdots & w_{NN_{l}} \\
  \end{array} } \right],
  h^{*} \in \mathbb{R}^{ N \times N_{l}}
  \label{eq:eq5}
\end{equation}
where $w$ means the label word.

\subsection{Population Evaluation}

\begin{algorithm}[t]
{\bf Input:}{ $S^{0}$; $D_{dev}$; $f$; $M$; $N_{iter}$; $P_{m}$; $P_{c}$\; }
{\bf Output:}{ The best optimized individual: $X^{*}$ \; }
  Initialize population $S^{0}=\left\{ X_{1},...,X_{M}\right\}$; $n=0$\;
  
  \While{$n<N_{iter}$}{
    {compute score of $X_{i} \in S^{n}$ using $f$ and remain top $M$ individuals\;}
    {Initialize a new empty population $S_{new}= \{ \}$; $m = 0$\;}
    \While{$m<M$}{
        {select two individuals $X_{i}$, $X_{j}$ from $S^{n}$ \;}
        \If{random(0,1) $<P_{c}$}{
        $X_{i,j}^{C}$ = Crossover($X_{i}$, $X_{j}$)\;
        \eIf{random(0,1) $<P_{m}$}
        {$X_{i,j}^{M}$ = Mutation($X_{i,j}^{C}$) and add $X_{i,j}^{M}$ in $S_{new}$\;}{add $X_{i,j}^{C}$ in $S_{new}$\;}
        }
        $m=m+1$\;
    }
    {$S^{n+1}$ = $S^{n}$ + $S_{new}$\;}
    {$n=n+1$\;}
  }
  {from $S^{N} = \{X_{*}^{1},...,X_{*}^{N}\}$, select the best individual $X^{*}$\;}
  \caption{Evolutionary Procedure}
  \label{alg:algorithm1}
 \Return{$X^{*}$\;}
\end{algorithm}
We formalize $X_{k} \in \mathbb{R}^{ N_{c}\times N_{c}}, k \in \{1,M\}$, as an individual in population, $M$ is the population size. To conduct the environmental selection, all members of the search population are evaluated. Our aim is to obtain the best $X^{*}$ during the evolutionary procedure. The procedure is described in \textbf{algorithm} \ref{alg:algorithm1}.

 In the initialization of the evolutionary procedure, $S^{0}$ is an initialized population and $f$ is the fitness function to compute the score of each individual $X_{k}$ and rank by the score to decide which individual will be reserved or eliminated at each iteration. Specifically, we use $X_{k}$ to get verbalizer by Eq.\ref{eq:eq4} and Eq.\ref{eq:eq5} and subsequently evaluate on the validation set to compute accuracy as fitness score. $N_{iter}$ is the maximum number of iterations. $P_{m}$ and $P_{c}$ are mutation probability and crossover probability, respectively. 
 
 The core of the evolutionary process is to reproduce the current generation of individuals and use the fitness function to select elite individuals iteratively. For each iteration, we firstly compute scores of $X_{k} \in S^{n}$ by using $f$ and remain top $M$ individuals. Then, we create a new empty set to store the generation of individuals. In the process of generation, we use the Roulette Wheel Selection algorithm~\cite{blickle1996comparison} to select two individuals from $S^{n}$ and generate new individuals by crossover and mutation with probability. The new generation will be added in $S^{n}$ and continue to iterate until the population has doubled in size.

After several steps of evolutionary search, we will collect the optimal individual $X^{*}$ from the final population and use Eq.\ref{eq:eq4} and Eq.\ref{eq:eq5} to obtain the final verbalizer. We show an illustration in Fig.\ref{fig:fig2}.

\subsection{Verbalizer Generation Strategies}
In the evolutionary verbalizer search, new verbalizers are generated in each generation, and the following verbalizer generation strategies are designed.
\paragraph{\textbf{Crossover operator}}
The Crossover operator is the process by which two individuals exchange some of their genes with each other in some way based on the crossover probability. The crossover operation plays a key role in EA and is the main method for generating new individuals. In the process of biological evolution, the probability of crossover is higher. In EVS, $P_{c}$ is set to 0.8. We randomly swap the elements of two rows of the matrix $X_{k}$ to obtain a new individual.
\paragraph{\textbf{Mutation operator}}
A mutation operator is the replacement of some gene values in an individual with other gene values based on the probability of mutation, resulting in a new individual.
$P_{m}$ is set to 0.1. We randomly initialize a matrix $X_{m}$ with the same shape as $X_{k}$ and these two matrices are multiplied together to obtain a new individual.

\section{Experiments}

The optimal result of EVS is \textbf{EvoVerb}. We conducted our experiments on five text classification datasets under few-shot settings to demonstrate the effectiveness of \textbf{EvoVerb}.
In this section, we first describe the dataset and experimental settings. Then, we analyze and discuss the experimental results.
\subsection{Experimental Settings}

\paragraph{\textbf{Dataset}} We conduct our experiment on five text classification benchmark datasets, including three topic classification datasets (AG`s News~\cite{zhang2015character}, DBPedia~\cite{lehmann2015dbpedia}, and Yahoo~\cite{zhang2015character}) and two sentiment classification datasets (IMDB~\cite{maas2011learning} and Amazon~\cite{mcauley2013hidden}). For Amazon, we randomly select 10000 instances from the original test set to construct a new test set. The statistics of each dataset are shown in Table~\ref{tab:dataset}.
\begin{table}[h]
    \centering
    \caption{The statistics of each dataset.}
    \begin{tabular}{c|cccc}
    \toprule
    Dataset & Type & \#Class & \#Training Size & \#Test Size \\
    \midrule
     AG’s News & Topic & 4 & 120000 & 7600 \\
     DBPedia & Topic & 14 & 560000 & 70000 \\
     Yahoo & Topic & 10 & 1400000 & 60000 \\
     Amazon & Sentiment & 2 & 3600000 & 10000 \\
     IMDB & Sentiment & 2 & 25000 & 25000 \\
    \bottomrule
    \end{tabular}
    \label{tab:dataset}
\end{table}
\paragraph{\textbf{Template}}
Due to the rich prior knowledge of human-picked templates, manual templates are better than auto-generated templates in few-shot settings. Following the previous works~\cite{hu2021knowledgeable}, we use the same four manual templates for each dataset. We test all baselines and our method by using four manual templates, and report both the average results (with standard error) of the four templates and the results of the best template.

\paragraph{\textbf{Evaluation}}

We use accuracy as micro-F1 for our experiments. We repeat 5 times with 5 different random seeds. 

\subsection{Baselines}

To verify the effectiveness of our method, we compare our method under few-shot setting with various approaches: fine-tuning, prompt-tuning with various state-of-the-art verbalizers. Fine-tuning, which uses the last layer`s hidden state of {\tt [CLS]} to a classification layer to make predictions. For prompt-tuning, we compare with: (1) \textbf{One2OneVerb}, using the class name as the only one label word for each class; (2) \textbf{One2ManyVerb}, using multiple label words for each class; (3) \textbf{KnowVerb}\cite{hu2021knowledgeable}, which selects related words from the large knowledge base to construct One2ManyVerb; (4) \textbf{AutoVerb}\cite{schick2020automatically}, searching label words from vocabulary automatically, which maximizes the likelihood of the training data; (5) \textbf{SoftVerb}\cite{hambardzumyan2021warp}, which treats the label words as trainable tokens and optimizes it with cross-entropy loss in the training processing; (6) \textbf{ProtoVerb}\cite{cui2022prototypical} learns prototype vectors as verbalizers by contrastive learning. The prototypes summarize training instances and can include rich label-level semantics.

\begin{table}[t]
    \centering
    \caption{Micro-F1 and standard deviation on five text classification datasets. \textbf{Bold}: best results in brackets and the best results among all methods for the same $k$-shot experiment.} 
    
    \scalebox{0.7}{\begin{tabular}{c|c|ccccc}
    \toprule
     $K$ & Verbalizer & AG`s News & DBPedia & Yahoo & Amazon & IMDB \\
    \midrule
        \multirow{7}*{1 shot} 
        & Fine-tuning & 29.5{\scriptsize$\pm$ 5.4} (35.3) & 26.6{\scriptsize$\pm$ 1.6} (31.3) & 13.1{\scriptsize$\pm$ 3.3} (16.6) & 45.5{\scriptsize$\pm$ 4.2} (50.4) & 51.4{\scriptsize$\pm$ 2.4} (54.6)  \\
        & One2OneVerb & 71.9{\scriptsize$\pm$ 7.2} (81.4) & 87.0{\scriptsize$\pm$ 1.0} (88.4) & 49.3{\scriptsize$\pm$ 1.2} (51.4) & 89.1{\scriptsize$\pm$ 3.1} (92.1) & 86.0{\scriptsize$\pm$ 3.3} (89.7) \\
        & One2ManyVerb & 74.3{\scriptsize$\pm$ 7.6} (\textbf{82.8}) & 89.8{\scriptsize$\pm$ 2.5} (92.3) & 50.9{\scriptsize$\pm$ 1.8} (52.6) & 88.7{\scriptsize$\pm$ 3.0} (91.6) & 88.4{\scriptsize$\pm$ 6.4} (\textbf{93.0}) \\
        & KnowVerb & \textbf{75.7}{\scriptsize$\pm$ 3.6} (78.4) & \textbf{90.3}{\scriptsize$\pm$ 3.7} (\textbf{93.6}) & \textbf{61.4}{\scriptsize$\pm$ 3.6} (\textbf{63.5}) & \textbf{89.4}{\scriptsize$\pm$ 4.2} (\textbf{92.8}) & \textbf{91.4}{\scriptsize$\pm$ 2.4} (92.7) \\
        & AutoVerb & 55.9{\scriptsize$\pm$ 2.8} (60.4) & 77.1{\scriptsize$\pm$ 1.7} (79.5) & 22.9{\scriptsize$\pm$ 4.2} (26.8) & 65.7{\scriptsize$\pm$ 9.3} (73.2) & 72.4{\scriptsize$\pm$ 7.1} (79.9) \\ 
        & SoftVerb & 65.1{\scriptsize$\pm$ 3.2} (68.8) & 62.9{\scriptsize$\pm$ 1.7} (65.7) & 32.5{\scriptsize$\pm$ 5.1} (37.2) & 71.7{\scriptsize$\pm$ 2.5} (74.2) & 79.1{\scriptsize$\pm$ 3.4} (82.2) \\
        & ProtoVerb & 64.3{\scriptsize$\pm$ 5.1} (72.4) & 74.9{\scriptsize$\pm$ 4.0} (79.4) & 48.3{\scriptsize$\pm$ 3.6} (51.4) & 69.6{\scriptsize$\pm$ 4.7} (73.6) & 79.3{\scriptsize$\pm$ 3.0} (83.5) \\
        & EvoVerb (ours) & 69.4{\scriptsize$\pm$ 3.4} (73.1) & 72.8{\scriptsize$\pm$ 1.4} (74.1)  & 44.4{\scriptsize$\pm$ 3.2} (47.7) & 73.4{\scriptsize$\pm$ 1.1} (73.9) & 72.7{\scriptsize$\pm$ 1.5} (73.6) \\
    \midrule
        \multirow{7}*{4 shot} 
        & Fine-tuning & 46.9{\scriptsize$\pm$ 3.2} (48.5) & 91.6{\scriptsize$\pm$ 1.2} (92.3) & 43.1{\scriptsize$\pm$ 8.2} (51.6) & 53.5{\scriptsize$\pm$ 3.3} (56.4) & 56.4{\scriptsize$\pm$ 1.4} (57.3)  \\
        & One2OneVerb & 77.9{\scriptsize$\pm$ 1.9} (79.4) & 94.2{\scriptsize$\pm$ 1.0} (95.8) & \textbf{62.8}{\scriptsize$\pm$ 1.2} (64.7) & 90.4{\scriptsize$\pm$ 1.8} (92.5) & 87.4{\scriptsize$\pm$ 7.2} (91.9) \\
        & One2ManyVerb & 78.5{\scriptsize$\pm$ 4.2} (83.5) & 94.3{\scriptsize$\pm$ 1.2} (96.0) & 62.5{\scriptsize$\pm$ 2.6} (\textbf{65.5}) & 93.1{\scriptsize$\pm$ 2.5} (93.6) & 91.2{\scriptsize$\pm$ 1.1} (\textbf{93.5}) \\
        & KnowVerb & 80.4{\scriptsize$\pm$ 5.6} (\textbf{84.8}) & 93.8{\scriptsize$\pm$ 2.5} (95.3) & 61.9{\scriptsize$\pm$ 4.8} (64.6) & 93.2{\scriptsize$\pm$ 1.2} (\textbf{94.1}) & 91.8{\scriptsize$\pm$ 0.6} (92.2) \\
        & AutoVerb & 73.1{\scriptsize$\pm$ 1.8} (74.8) & 92.4{\scriptsize$\pm$ 1.3} (94.0) & 57.7{\scriptsize$\pm$ 3.3} (60.0) & 87.8{\scriptsize$\pm$ 3.1} (91.2) & 90.0{\scriptsize$\pm$ 3.7} (93.4) \\ 
        & SoftVerb & 76.9{\scriptsize$\pm$ 4.7} (81.5) & 93.8{\scriptsize$\pm$ 1.6} (96.5) & 48.4{\scriptsize$\pm$ 3.9} (54.7) & 92.4{\scriptsize$\pm$ 2.5} (93.7) & 87.1{\scriptsize$\pm$ 3.3} (90.6) \\
        & ProtoVerb & 79.6{\scriptsize$\pm$ 1.6} (82.3)  & \textbf{95.5}{\scriptsize$\pm$ 0.7} (96.2) & 60.5{\scriptsize$\pm$ 3.9} (64.6) & 90.2{\scriptsize$\pm$ 1.5} (92.0) & 91.5{\scriptsize$\pm$ 2.5} (92.2) \\
        & EvoVerb (ours) & \textbf{80.7}{\scriptsize$\pm$ 1.7} (82.1) & 94.6{\scriptsize$\pm$ 2.8} (\textbf{96.7})  & 57.2{\scriptsize$\pm$ 4.1} (60.9) & \textbf{93.4}{\scriptsize$\pm$ 1.4} (93.6) & \textbf{92.5}{\scriptsize$\pm$ 1.7} (93.3) \\
    \midrule
        \multirow{7}*{8 shot} 
        & Fine-tuning & 73.5{\scriptsize$\pm$ 4.4} (78.9) & 92.5{\scriptsize$\pm$ 0.5} (93.2) & 48.1{\scriptsize$\pm$ 2.5} (53.6) & 83.8{\scriptsize$\pm$ 2.3} (86.7) & 77.3{\scriptsize$\pm$ 3.7} (81.4)  \\
        & One2OneVerb & 82.4{\scriptsize$\pm$ 2.0} (84.7) & 95.9{\scriptsize$\pm$ 1.2} (97.2) & 64.5{\scriptsize$\pm$ 0.3} (64.9) & 90.5{\scriptsize$\pm$ 2.7} (92.2) & 91.3{\scriptsize$\pm$ 1.4} (93.6) \\
        & One2ManyVerb & 84.3{\scriptsize$\pm$ 1.9} (86.9) & 97.3{\scriptsize$\pm$ 0.7} (98.0) & 64.4{\scriptsize$\pm$ 0.8} (65.5) & 93.6{\scriptsize$\pm$ 0.6} (94.3) & 91.1{\scriptsize$\pm$ 3.6} (93.2) \\
        & KnowVerb & 83.3{\scriptsize$\pm$ 5.6} (86.8) & 89.8{\scriptsize$\pm$ 2.5} (92.3) & 66.9{\scriptsize$\pm$ 1.8} (68.6) & 93.7{\scriptsize$\pm$ 0.5} (94.2) & 92.1{\scriptsize$\pm$ 1.3} (92.7) \\
        & AutoVerb & 80.0{\scriptsize$\pm$ 4.7} (85.2) & 95.7{\scriptsize$\pm$ 2.0} (97.8) & 60.5{\scriptsize$\pm$ 1.6 (62.5)} & 88.5{\scriptsize$\pm$ 8.4} (94.1) & 91.6{\scriptsize$\pm$ 2.5} (92.9) \\ 
        & SoftVerb & 81.4{\scriptsize$\pm$ 1.9} (82.8) & 96.5{\scriptsize$\pm$ 0.3} (97.0) & 51.5{\scriptsize$\pm$ 6.5 (58.0)} & 91.0{\scriptsize$\pm$ 3.1} (93.0) & 91.6{\scriptsize$\pm$ 0.7} (92.3) \\
        & ProtoVerb & 85.4{\scriptsize$\pm$ 1.9} (87.9) & 97.4{\scriptsize$\pm$ 0.6} (\textbf{98.2}) & 66.7{\scriptsize$\pm$ 1.2} (68.0) & 92.4{\scriptsize$\pm$ 1.8} (94.4) & 89.1{\scriptsize$\pm$ 5.0} (92.6) \\
        & EvoVerb (ours) & \textbf{86.5}{\scriptsize$\pm$ 1.5} (\textbf{88.1}) & \textbf{97.8}{\scriptsize$\pm$ 0.3} (98.1)  & \textbf{69.4}{\scriptsize$\pm$ 0.7} (\textbf{69.9}) & \textbf{94.0}{\scriptsize$\pm$ 1.1} (\textbf{94.5}) & \textbf{92.7}{\scriptsize$\pm$ 1.6} (\textbf{93.7}) \\
    \midrule
        \multirow{7}*{16 shot} 
        & Fine-tuning & 84.1{\scriptsize$\pm$ 3.5} (87.1) & 95.6{\scriptsize$\pm$ 1.2} (96.3) & 57.1{\scriptsize$\pm$ 2.7} (60.5) & 85.3{\scriptsize$\pm$ 2.1} (88.6) & 76.4{\scriptsize$\pm$ 1.4} (77.3)  \\
        & One2OneVerb & 85.5{\scriptsize$\pm$ 1.6} (88.3) & 97.5{\scriptsize$\pm$ 0.4} (98.2) & 66.5{\scriptsize$\pm$ 1.7} (69.2) & \textbf{93.1}{\scriptsize$\pm$ 0.7} (94.0) & 92.1{\scriptsize$\pm$ 1.0} (92.8) \\
        & One2ManyVerb & 86.9{\scriptsize$\pm$ 0.8} (87.7) & 97.3{\scriptsize$\pm$ 0.6} (98.0) & 65.9{\scriptsize$\pm$ 3.1} (69.1) & 90.6{\scriptsize$\pm$ 3.3} (94.4) & 92.3{\scriptsize$\pm$ 5.2} (\textbf{93.9}) \\
        & KnowVerb & 85.2{\scriptsize$\pm$ 4.6} (86.8) & 97.8{\scriptsize$\pm$ 1.5} (97.8) & 68.1{\scriptsize$\pm$ 0.9} (68.6) & 91.7{\scriptsize$\pm$ 2.4} (93.6) & 91.4{\scriptsize$\pm$ 1.4} (92.7) \\
        & AutoVerb & 86.6{\scriptsize$\pm$ 2.7} (88.0) & 96.2{\scriptsize$\pm$ 2.0} (97.7) & 64.4{\scriptsize$\pm$ 1.8} (67.5) & 92.9{\scriptsize$\pm$ 1.7} (93.9) & 91.5{\scriptsize$\pm$ 2.4} (93.1) \\ 
        & SoftVerb & 85.6{\scriptsize$\pm$ 0.9} (86.6) & 97.8{\scriptsize$\pm$ 0.1} (98.0) & 62.8{\scriptsize$\pm$ 1.8} (64.5) & 91.0{\scriptsize$\pm$ 3.1} (94.2) & 90.4{\scriptsize$\pm$ 4.4} (93.7) \\
        & ProtoVerb & 85.7{\scriptsize$\pm$ 1.6} (87.3) & 97.3{\scriptsize$\pm$ 0.7} (98.2) & 69.0{\scriptsize$\pm$ 1.1} (70.2) & 92.5{\scriptsize$\pm$ 1.7} (\textbf{94.5}) & 89.7{\scriptsize$\pm$ 1.6} (92.6) \\
        & EvoVerb (ours) & \textbf{87.2}{\scriptsize$\pm$ 0.6} (\textbf{88.6}) & \textbf{98.0}{\scriptsize$\pm$ 0.4} (\textbf{98.3})  & \textbf{70.3}{\scriptsize$\pm$ 0.7} (\textbf{70.7}) & 
        93.4{\scriptsize$\pm$ 2.3} (93.7) & \textbf{93.2}{\scriptsize$\pm$ 1.4} (93.5) \\
    \bottomrule
    \end{tabular}}
    \label{tab:few-result}
\end{table}

\subsection{Implementation Details}

All our experiments are based on Transformers\footnote{\url{https://github.com/huggingface/transformers}} and OpenPrompt~\cite{ding2021openprompt} framework. We employ RoBERTa-large~\cite{liu2019roberta} as our pre-trained language model backbone and use AdamW optimizer with learning rate 3e-5 to optimize it. For few-shot setting, we conduct 1,4,8,16-shot experiments. For a $k$-$shot$ experiment, we sample $k$ instances of each class from the original training set to form the few-shot training set and sample another $k$ instances per class to form the validation set. The max sequence length is set to 256, and we train the model for 5 epochs with the batch size set to 8 and choose the best checkpoint in validation set for each experiment. In EVS, the validation set is also used as the development set to search verbalizers. The size of population is set to 30 and the maximum iteration is set to 5. $N_{c}$ and $N_{l}$ are set to 1000 and 100 respectively. We run all experiments using PyTorch 1.7.0 on a single Nvidia RTX 3090 GPU. Our code is available at \url{https://github.com/rickltt/evs}.

\subsection{Results}

In this part, we analyze the results of few-shot experiments with different verbalizers and provide insights of EvoVerb. 

From Table \ref{tab:few-result}, we find out that all prompt-based methods outperform fine-tuning methods under five datasets. The gap is decreasing as the shot increases. Compared to other baseline verbalizers, One2ManyVerb performs slightly better than One2OneVerb and KnowVerb wins over One2ManyVerb by a slight margin. The result proves that one-many mapping can considerably enhance the probability of making correct predictions and one-one mapping lacks enough information for predictions. KnowVerb introduces an external knowledge base to enrich the coverage of label words, so it can yield better performance. Although free of manual design, AutoVerb, SoftVerb and ProtoVerb have poor performance under data scarcity. As the training data increase sufficiently, they get even exceeding scores. The reason is obvious, that these methods require enough data to optimize during the training process. 

When comparing EvoVerb with the other optimize-based verbalizers, we find EvoVerb outperforms other baseline methods on 8 and 16 shot and achieves similar performance on 4-shot. But it is quite unsatisfactory on 1-shot. We attribute the reason why our methods do not perform as well as other verbalizers in 1-shot and 4-shot scenarios to too few training and development samples, which result in inadequate initialization for evolutionary verbalizer search. When in 8-shot and 16-shot settings, our method achieves better performance in prompt-tuning, which indicates that more instances can result in a better representation of the initialization labels and promote EVS to search high performance verbalizer.

\subsection{Analysis}

In this subsection, we discuss several factors that affect EVS.
\paragraph{\textbf{The Size of Development Set}} 

In EVS, the development set plays an important role in the initialization of each label representation and is also critical to the final high-performance verbalizer from EVS. To investigate the best searched verbalizer with different size of development set, we present $k$-shot experiments for 1 and 16 on AG's News dataset. From Table.~\ref{tab:labelword}, we see that: (1) Even if there is only one example per class, the searched verbalizer is meaningful. (2) With more examples, EVS can find more appropriate label words. Most similar label words are closely related to the corresponding label topics.
\begin{table}[t]
    \centering
    \caption{The result of label words for each label on AG's News dataset obtained using EVS. Select top 5 label words for each label.}
    \scalebox{0.8}{\begin{tabular}{l|c|c}
    \toprule
    Label & $K$ = 1 & $K$ = 16 \\
    \midrule
     World & headlines, news, report, foreign, country & governance, politics, world, government, national  \\
     Sports & score, match, playoff, replay, repeat & racing, sprinter, netball, sporting, athletics   \\
     Business & chemical, cbs, systemic,committee , auditor & company, bank, corruption, billboard, taxpayer \\
     Tech & device, security, digital, phone, energy & facebook, technical, computer, internet, nanotechnology \\
    \bottomrule
    \end{tabular}
    }
    \label{tab:labelword}
\end{table}
\begin{figure}[t]
	\centering
	\begin{minipage}{0.49\linewidth}
		\centering
		\includegraphics[width=1.0\linewidth]{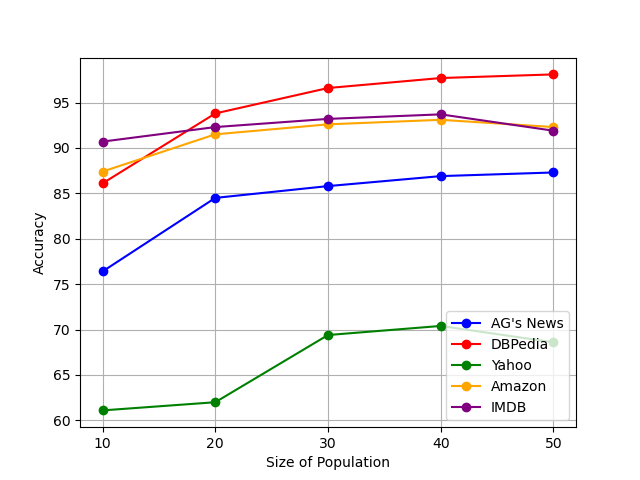}
		\caption{Ablation results on the size of population.}
		\label{popsize}
	\end{minipage}
	\begin{minipage}{0.49\linewidth}
		\centering
		\includegraphics[width=1.0\linewidth]{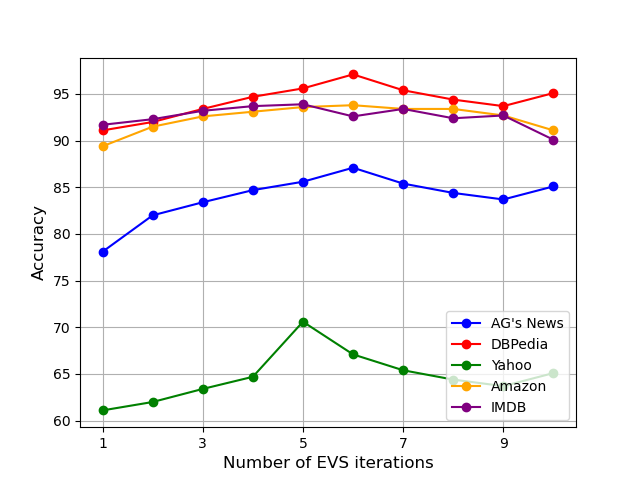}
		\caption{Ablation results on the number of EVS iterations.}
		\label{iters}
	\end{minipage}
\end{figure}

\paragraph{\textbf{Population Size}} 

The population size is also a critical hyperparameter for EVS. As shown in Fig.~\ref{popsize}, the overall performance on all datasets improves first and then declines with population growth, which indicates that large population sizes do not optimize the results of the evolutionary algorithm.

\paragraph{\textbf{The Number of EVS Iterations}}

Another critical hyperparameter is the number of iterations for the evolutionary verbalizer search. We conduct experiments from 1 to 10 iterations on five datasets and the results are shown in Fig~.\ref{iters}. It can be seen that some datasets such as Yahoo and IMDB achieve the best results in iteration 5, but other datasets achieve the best results in iteration 6. EVS requires enough iterations to find the high-performance verbalizers. In order to balance time cost and performance, the default iteration is set to 5 in EVS.

\section{Conclusion}

In this paper, we propose a novel approach for automatic verbalizer construction in prompt-based tuning. To obtain a high-performance and meaningful verbalizer, we optimize models by utilizing evolutionary algorithm. We compare it to other current verbalizers and the experimental results demonstrate that EvoVerb outperforms various baselines and verifies its effectiveness. For future work, we will focus on extending EvoVerb for effective strategies of evolutionary algorithm. 

\subsubsection{Acknowledgements} This work was supported in part by the National Natural Science Foundation of China (62006044, 62172110), in
part by the Natural Science Foundation of Guangdong Province (2022A1515010130), and in part by the Programme of Science and Technology of Guangdong Province (2021A0505110004).

%
%
%
\bibliographystyle{splncs04}
\bibliography{reference}

\end{document}